\documentclass[pdflatex,sn-mathphys-num]{sn-jnl}

\usepackage{graphicx}%
\usepackage{multirow}%
\usepackage{amsmath,amssymb,amsfonts}%
\usepackage{amsthm}%
\usepackage{mathrsfs}%
\usepackage[title]{appendix}%
\usepackage{xcolor}%
\usepackage{textcomp}%
\usepackage{manyfoot}%
\usepackage{booktabs}%
\usepackage{algorithm}%
\usepackage{algorithmicx}%
\usepackage{algpseudocode}%
\usepackage{listings}%
\usepackage{comment}%

\theoremstyle{thmstyleone}%
\theoremstyle{thmstyletwo}%

\theoremstyle{thmstylethree}%

\begin{document}

\title[Article Title]{Scaling Science: Toward a Chain Reaction of Discovery}


\author*[1]{\fnm{Yuheng} \sur{Ding}}\email{ydingcd@connect.ust.hk}

\affil*[1]{%
    \orgdiv{Department of Electronic and Computer Engineering},
    \orgname{The Hong Kong University of Science and Technology},
    \orgaddress{\city{Hong Kong}, \country{China}}}


\abstract{Scientific discovery advances when evidence from different investigations is brought into contact. Yet paper-centric communication leaves experimental results fragmented, delayed, or unreported, creating a ``scientific memory wall'' between growing computational capabilities and the evidence accessible to AI systems. Breaking this wall requires a paradigm shift in AI for Science: moving from passively mining questions out of existing datasets to actively generating and recombining experimental data around unresolved scientific questions. In a network of intelligent laboratories, a failed or overlooked experiment in one domain could trigger a hypothesis in another, whose testing generates fresh evidence that stimulates further investigations. To enable these evidence-driven chain reactions, we propose Evolving Scientific Intelligence (ESI), an architectural blueprint that connects shared scientific memory with distributed experimental loops. ESI preserves experimental records—including negative and anomalous results—in computable representations, allowing models to recombine evidence across domains into testable hypotheses. Rather than following fixed screening pipelines, the system jointly selects candidates and evaluation methods according to scientific value, reliability, and cost, directing distributed laboratories to produce the exact evidence needed next. Their physical feedback revises both hypotheses and evaluators, while slower meta-control reassesses research objectives and decision criteria to guard against self-reinforcing errors and misguided optimization. We call for an Information Manhattan Project to build this infrastructure for scaling science through experimentally grounded chains of discovery.}

\maketitle

\section{Introduction}\label{sec1}

In the ancient myth of Babel, human ambition converged on the plains of Mesopotamia to construct a tower toward the heavens—a testament to an unbounded potential driven by a shared language. When that common tongue dissolved, the tower was left as a silent monument to divided purpose. Modern science faces a comparable problem: its expanding body of knowledge is divided across disciplines and recorded in narrative prose or incompatible representations. These barriers hinder the synthesis of scientific knowledge by AI systems and the coordination of experiments across automated laboratories.

Whether within an individual mind or across individuals, our understanding of the world fundamentally emerges from the collision of information. Experiments and computational simulations, together with theoretical reasoning, produce representations of physical reality—a process we call information production. These representations become the material for information collision, in which evidence from different sources is connected and structurally recombined to generate hypotheses. Testing these hypotheses produces fresh evidence, opening further possibilities for discovery. The result is a continuous cycle in which knowledge changes our understanding of the world.

For centuries, this cycle has relied on researchers bringing evidence reported in scientific papers into contact with their own knowledge and experience. Independently published findings can contain latent connections from which new discoveries emerge~\citep{swanson1986undiscovered}. Large language models (LLMs) now extend this process of knowledge recombination~\citep{ghafarollahi2025sciagents}, while research papers remain the gold standard for synthesizing findings and assigning scholarly credit.

Even so, papers alone cannot carry the full stream of experimental evidence needed to train models and guide automated laboratories. In constructing coherent narratives, researchers often leave negative results and exploratory trials, as well as unexplained observations, outside the published record. Much of this information remains inaccessible or difficult to reuse as `dark data'~\citep{franco2014publication,raccuglia2016machine}. Positive results supporting the main hypotheses, meanwhile, are more likely to be reported. This selective reporting introduces systematic data bias, narrowing the empirical evidence base while limiting what models can learn about the conditions under which experiments succeed or fail~\citep{strieth2022machine}. Moreover, even when valuable experimental evidence is eventually published, the time required for it to reach and be utilized by relevant researchers may far exceed the time needed to produce it. Amid the growing volume of scientific literature, such evidence may also be overlooked by precisely those researchers whose knowledge could reveal its latent scientific value.

This constraint resembles the memory wall in computer architecture, where processing performance grows far faster than memory bandwidth~\citep{wulf1995hitting}. Scientific AI faces an analogous bottleneck: publication delays and selective reporting restrict access to evidence, while the narrative-heavy presentation of experimental evidence in scientific papers limits its computational accessibility and reuse by increasingly capable AI systems~\citep{franco2014publication,jablonka2022making}. We term this mismatch between computational capability and the availability of reliable evidence the ``scientific memory wall''. Overcoming this barrier requires a shift from passively mining existing datasets to actively generating and preserving experimental evidence around unresolved scientific questions—complementing formal publication, which credits contributors while providing a traceable record of the research.

Scaling AI for Science requires more than scaling models and datasets. It also requires the ability to systematically guide information collisions, enabling discoveries to propagate through evidence-driven chain reactions. To make such collisions possible, scientific information must be organized into computable representations that preserve experimental context while exposing shared information interfaces across fields. Because an unexplained anomaly or negative trial may acquire new meaning as research questions shift or new evidence arrives, current models should not discard records solely on the basis of their immediate utility. Through these interfaces, LLMs can recombine evidence from separate investigations to generate testable hypotheses that neither study originally set out to answer.

Yet as shared memory and information collision make hypothesis generation increasingly rapid, the primary bottleneck of discovery shifts to evaluation and physical feedback. Before the advent of LLMs, methods such as Bayesian optimization (BO) and active learning (AL) guided experimental selection through predefined acquisition strategies~\citep{shields2021bayesian,hennig2012entropy}. More recent evolutionary frameworks such as FunSearch and AlphaEvolve draw on learned knowledge and evaluator feedback to iteratively propose new code candidates~\citep{novikov2025alphaevolve,romera2024mathematical}. Extending this evolutionary loop to the physical and life sciences is far more challenging: high-fidelity experiments are costly and slow, while computational predictors—such as structure predictors used to screen protein designs—still struggle with flexible regions like antibody complementarity-determining regions (CDRs) and can introduce false positives or reinforce their own biases if left unchecked~\citep{woo2024protein,pacesa2025one}. Rather than relying on fixed screening pipelines, scientific intelligence must treat evaluation itself as a decision—jointly selecting candidates and evaluation methods while using independent physical measurements to recalibrate the evaluators.

Closing this loop at scale requires moving beyond isolated automated laboratories. While LLMs can already design experiments for individual robotic platforms to execute~\citep{boiko2023autonomous,darvish2025organa,smith2026using}, standalone loops remain bounded by local hardware, fixed background conditions, and the risk of compounding setup-specific biases across iterations. Connecting heterogeneous laboratories through shared representations and capability interfaces allows evidence produced at one node to trigger hypotheses and targeted verification at others, turning isolated workflows into experimentally grounded chain reactions of discovery.

To sustain and govern these chain reactions, we propose Evolving Scientific Intelligence (ESI), an architectural blueprint for scaling science across distributed experimental loops. ESI couples shared, computable scientific memory with a three-level control hierarchy: L1 closes the fast sensing-and-execution loop at local laboratory nodes; L2 coordinates cross-domain information collision, joint candidate--evaluator selection, and asynchronous task dispatch across laboratories; and L3 operates on a slower timescale, auditing cross-node provenance to revise control rewards, evaluation criteria, and research objectives when proxy metrics diverge from genuine progress. Together, these mechanisms offer a path toward an Information Manhattan Project—an infrastructure in which accumulated evidence continuously reshapes both what science knows and how it discovers.

\section{The Scientific Memory Wall}\label{sec2}

Scaling scientific foundation models without reforming how experimental evidence is captured and shared merely amplifies a fundamental bottleneck: the scientific memory wall. As computational capacity continues to grow, access to high-quality data is becoming an increasingly important bottleneck for further advances in AI. A research system can face data scarcity while also wasting information it has already produced. Meanwhile, the dissemination of experimental data through research papers struggles to keep pace with growing demand. Data already obtained may be excluded from the scientific record because they do not support the expected conclusions. Beyond wasting time and resources, selective reporting changes the distribution of data available to models and can impair their generalization and predictive performance~\citep{strieth2022machine}. Delayed access is equally consequential. Even when experimental results eventually enter public datasets or appear in papers, substantial time may pass between their generation and release~\citep{ioannidis1998effect}. Until then, other researchers and models cannot use them for further analysis or experimental decisions. Clearly, the value of information also depends on when it becomes available. Viewed this way, the scientific memory wall concerns both the production of reliable experimental data and what happens to those data afterward. Addressing it requires producing more evidence and preserving it for reuse. Furthermore, reducing delays in validation and data sharing would allow each experiment to inform subsequent work sooner.

\subsection{The Missing Ground Truth}

In the context of the scientific memory wall, biomolecular research provides a clear example of the data shortage. Advances in protein structure prediction allow structures to be predicted for more sequences, yet the modeling of flexible regions in many biomolecules remains constrained by limited experimental data. For example, antibody CDRs vary considerably in length and amino acid composition, particularly the third CDR of the heavy chain (CDR H3), making their structures difficult to model accurately. Woo et al. noted that antibody loops are underrepresented in existing training resources, limiting prediction accuracy~\citep{woo2024protein}. At the same time, sequence data do not necessarily come with functional measurements. When developing an antibody binding dataset, Engelhart et al. noted that many sequences in large antibody resources lacked corresponding target and binding affinity information~\citep{engelhart2022dataset}. Learning is therefore also limited by missing connections between different kinds of information: which conformation a sequence adopts, under what conditions it remains stable, and whether it binds to a particular target. Connecting these observations is essential for moving from structure prediction to functional design.

These data gaps can propagate through the dependence of generative methods on structure predictors. In protein design, structure predictors are widely used to generate and screen candidates. Hallucination-based methods like BindCraft incorporate structure prediction models such as AlphaFold2 into the design process and use them to score and filter candidate designs~\citep{pacesa2025one}. Depending on the optimization objective, the design process may become tuned to the predictor’s scoring preferences or depend heavily on its structure prediction accuracy. Limitations in the predictor can also affect screening after generation. Higher computational scores therefore do not necessarily translate into corresponding improvements in experimental performance. The BindCraft study also reported a substantial proportion of false-positive predictions when AlphaFold3 was used to screen candidates~\citep{pacesa2025one}. These findings highlight the need to test high-scoring candidates experimentally and feed discrepancies between predictions and measurements back to the model to guide the next round of design.

\subsection{The Dark Data of Science}

Some data gaps arise because experiments have not yet been performed; others emerge after experiments are complete. Papers organize evidence around a particular question. Results that do not support the main hypothesis, intermediate records from exploratory work and observations that remain unexplained often fall outside the final narrative. By tracking completed social science experiments, Franco et al. found that study outcomes strongly influenced whether findings were written up and published; many null results were never submitted~\citep{franco2014publication}. In chemical reaction prediction, Strieth-Kalthoff et al. examined biases in experiment selection and result reporting, showing how the omission of negative examples affects the learning of chemical reactivity~\citep{strieth2022machine}. Together, these studies reveal a common mechanism: when records are retained based on their outcomes, models learn from an outcome-filtered experimental distribution.

This mechanism explains why negative results are valuable. Most directly, they help establish the conditions under which an experimental approach succeeds. Comparing successful and unsuccessful trials can reveal limits that are difficult to identify from successful cases alone. In hydrothermal materials synthesis, a model trained on historical reaction records, including failed experiments, recommended conditions that achieved an 89\% success rate, compared with 78\% for conditions selected through human expertise~\citep{raccuglia2016machine}.

Differences between unsuccessful experiments can also provide clues about how to improve experimental outcomes. In microfluidic particle separation, images of unsuccessful trials supplied graded feedback for reinforcement learning, helping a controller identify effective operating conditions~\citep{fukada2022microfluidic}. In experimental optimization, GOLLuM began each task with ten low-performing observations and used successive rounds of sampling to identify high-performing candidates, ranking first on average across 23 benchmarks~\citep{rankovic2026large}. These examples show how unsuccessful or low-performing experiments can inform subsequent decisions.

The value of dark data also extends beyond success and failure. A measurement unrelated to one paper’s central question may provide precisely the information needed for another research question. An unexplained anomaly may acquire meaning when considered alongside other observations. Data retention should therefore preserve observations whose relevance is not yet apparent, allowing their value to emerge through later analysis and reuse.

It is also essential to note that data retention and timely access are separate requirements but equally important. Even data that will eventually be released may remain unavailable to external models while a paper is being prepared. In this sense, data awaiting publication can also be considered temporarily dark: although already produced, they remain invisible to the broader scientific community and may miss the window in which they would be most valuable. During this interval, other laboratories may repeat unsuccessful experiments or miss evidence that could change their experimental plans. In an iterative research system, delayed release extends the learning cycle; reducing this delay should be a design objective for scientific information infrastructure.

\subsection{From Passive Archiving to Automated High-Throughput Production}

Overcoming the scientific memory wall requires shifting from mining questions out of static datasets to generating experimental data around defined scientific questions. Tsuboyama et al. used cDNA display proteolysis to conduct large-scale measurements under consistent experimental conditions, producing approximately 776,000 protein folding stability measurements after quality filtering~\citep{tsuboyama2023mega}. Ribonanza assembled chemical probing measurements for approximately two million RNA sequences and used them for model training and prospective evaluation~\citep{he2024ribonanza}. These studies suggest that database development can extend from collecting existing results to actively organizing measurements around defined gaps in knowledge, with each experiment adding to a cumulative knowledge base.

Continuously generated data also need a common recording framework. The Protein Data Bank (PDB) provides a standardized archive where experimental structures from different institutions undergo validation and expert curation for shared analysis~\citep{burley2023rcsb}. Following this principle, databases for model learning should link each measurement to its sample, experimental conditions, procedure, and data source~\citep{kearnes2021open}. Recording these relationships consistently also reduces the work of reconstructing experimental context from prose, figures and supplementary materials.

Self-driving laboratories (SDLs) can sustain this process of data production. For example, SAMPLE links protein sequence design to biochemical characterization through gene assembly and expression. The results then guide the model’s next round of selection~\citep{rapp2024self}. Similarly, Clio refines electrolyte formulations through automated mixing and physical-property measurements, using the feedback to plan subsequent experiments~\citep{dave2022autonomous}. Active sampling further allows SDLs to direct measurements towards unresolved questions. For example, CAMEO integrates phase mapping and materials property optimization within the same experimental loop, allowing subsequent measurements to support both knowledge acquisition and the search for improved performance~\citep{kusne2020fly}. An experiment may therefore be selected for what it could reveal, even if it is not predicted to give the best result.

In the system we envisage, data capture and quality checks happen as part of the experiment itself rather than as an afterthought. Failed trials and unexpected anomalies are logged alongside successful outcomes, preserving the conditions under which each result was obtained. Once a record passes validation, it can immediately inform model updates and the next round of experimental planning without waiting to be packaged into a paper. Combining high-throughput execution with this kind of immediate, unfiltered recording turns experimental databases into a continuously growing scientific memory. Furthermore, reuse across platforms requires representations that preserve scientific meaning and support computational use.

\section{Beyond Accumulation: Records, Representations, and Synergy}\label{sec3}

The scientific memory wall reflects the gap between the growing data needs of scientific intelligence and the evidence it can access and use. Accumulating more scientific data, however, does not necessarily make scientific systems more intelligent. Moving from raw accumulation to genuine understanding requires ESI to answer three questions: Why should the retention of experimental records not be determined solely by their current uses? How can suitable representations reduce the computational cost of understanding and using evidence? And how can evidence from different sources be connected and recombined to generate new hypotheses and guide experiments?

\subsection{The Evolving Value of Experimental Records}

Today's models must not be the final arbiters of information's future scientific value. What appears uninformative for a present task may become essential when research questions change, new evidence emerges, or models gain new capabilities. ESI should therefore distinguish what information is useful to process now from what evidence deserves to remain available for future inquiry.

Assessing a record’s current contribution remains important. Data Shapley provides a formal approach to data valuation: given a learning algorithm and an evaluation metric, it measures the average marginal contribution of an individual training data point to predictive performance~\citep{ghorbani2019data}. However, these valuations are specific to a task and method. They can guide current data selection but may not reflect a record’s value in future research.

As new research questions emerge, existing records can serve new purposes. For example, a study may initially focus only on final yield, but later seek to understand reaction rates, product inhibition, or catalyst deactivation. If the original records preserve the full course of the reaction, researchers can reanalyze these processes to understand how the outcome arose. Previous studies have shown that analyzing changes in concentration over time can reveal kinetic information, such as reaction orders, and provide evidence for evaluating reaction mechanisms~\citep{blackmond2005reaction,bures2016variable}.

New evidence can also change the meaning of old records. For example, if an antibody variant produces no detectable target-binding signal, that result alone often cannot tell whether the variant binds weakly or instead suffers from low expression or poor folding stability. As measurements of these properties accumulate across additional variants, researchers can revisit the original observation and gradually separate these explanations~\citep{de2026effects}. Old records can thus help answer questions that could not initially be resolved.

Historical data may also yield new insights as analytical methods improve. Kozlov et al. used machine-learning-assisted methods to analyze approximately 22,000 historical mass spectrometry records, searching for previously unrecognized reactions and products, and validated some of these findings through further experiments~\citep{kozlov2025discovering}. 

For old records to support new analyses, the experimental conditions and procedures that produced the results must be preserved alongside the results themselves, providing a potential basis for future causal analysis and counterfactual reasoning. For example, researchers can compare outcomes across conditions and use an explicit causal model to ask what would have happened if only one condition had changed. Such analyses can clarify which conclusions the records support and where further experiments are needed~\citep{pearl2009causal}. Preserving this information is especially important when the original samples have been consumed or the experimental environment has changed.

ESI therefore needs to preserve experimental conditions, procedures, measurements, and data provenance as comprehensively as possible, allowing future researchers to trace and reanalyze experiments. Current valuations can guide which records are used first, but should not alone determine which records are preserved for future research.

However, the continued accumulation of these records increases the cost of storing and transmitting data, while also demanding more computation for processing. ESI therefore needs to curb duplicate copies and repeated descriptions without losing the original evidence. Suitable organization and encoding can further reduce the computational effort required to interpret these records for downstream use.

\subsection{A Language for Scientific Intelligence}

Beyond its reliance on rapid validation, AlphaEvolve~\citep{novikov2025alphaevolve} also owes its success to the way code expresses information: compared with natural language, code has a more explicit structure and less redundancy. Research papers, by contrast, are a narrative compression format optimized for human persuasion rather than machine reasoning; in science, making a record retrievable as text does not make its underlying evidence computationally usable~\citep{wilkinson2016fair}. Organizing experimental records into machine-readable structures allows models to access their content without repeatedly extracting it from prose. Incorporating relevant physical laws and operational constraints further helps systems assess whether findings are comparable and experimental procedures can be combined under specific conditions. More computation can then be devoted to learning regularities that inform hypothesis generation and experimental design.

The theory of usable information provides a foundation for this argument. $\mathcal{V}$-information measures the extent to which data can help a class of models improve its predictions. Researchers have shown that, because models have limited capabilities, more suitable representations can help them learn relationships that would otherwise be difficult to identify, even without additional observations, thereby making fuller use of existing information~\citep{xu2020theory}.

Natural language remains important for communicating scientific findings in a readable form that researchers can review and reassess. In the longer term, however, scientific information may not need to pass through natural language when exchanged between AI systems. Directly exchanging machine representations can substantially reduce the costs of conversion to natural language and repeated parsing. Previous work has provided initial evidence for the feasibility of this approach by transmitting internal model representations between agents~\citep{du2026enabling}. We envision that this mode of exchange will eventually become a reality, but it requires different models to share explicit encoding conventions and scientific semantics, a requirement that remains difficult to meet today.

Given the substantial heterogeneity of scientific data, a more practical direction at present is to share common semantic structures while allowing different types of data to use suitable encodings. For example, small molecules can be represented using SMILES, molecular graphs, or fingerprints, while proteins and nucleic acids can be represented by their sequences or learned embeddings. Associated experimental information can be organized into standardized fields, and physical laws and operational constraints can be expressed as computationally checkable rules.

Existing studies provide foundations for detailed implementation at several levels. Scientific entities scattered across papers can be organized into knowledge graphs, making them both retrievable and easier to combine~\citep{luan2018multi}. This approach can also be extended to experimental systems, where knowledge graphs connect materials, instruments, data, and agents to support information sharing and coordination~\citep{bai2022platform}.

Such representations can also support experimental execution and  enable the transformation of research papers into interactive AI agents. Previous studies have used machine-readable rules to describe when an operation can be performed and how it changes the system state. Agents use these rules to assemble experimental workflows and simulate the state after each operation to determine whether subsequent steps can be executed~\citep{li2026computable}. Paper2Agent transforms papers and their accompanying code into interactive AI agents. Researchers can describe their needs in natural language, and the agents can run the methods presented in the papers. In applications such as single-cell analysis, these agents can reproduce existing analyses and apply the methods to new research~\citep{miao2026reimagining}.

By preserving scientific meaning while reducing the computational burden of interpreting and integrating heterogeneous information, these representations enable models to reuse existing knowledge more effectively across research tasks. Scientific representations are therefore not merely formats for storing and exchanging data, but a fundamental determinant of how effectively scientific intelligence can learn from accumulated evidence.

\subsection{When Information Collides}

A counterintuitive observation is that the recombination of existing information can generate new understanding even without new observational data. The continued evolution of scientific intelligence fundamentally depends on whether evidence from multiple sources can be connected and recombined across domains. Even when experimental records are comprehensively preserved and consistently encoded, knowledge silos persist if their use remains confined to their original tasks and disciplines. ESI must therefore actively seek valuable combinations that make existing evidence useful for new research questions across domains and scales.

Partial Information Decomposition (PID) provides a theoretical language for understanding the value of combinations~\citep{williams2010nonnegative}. For two information sources, \(X_1\) and \(X_2\), and a shared prediction target, \(Y\), their joint mutual information can be decomposed into redundant information \(R\), which both sources provide; unique information \(U_1\) and \(U_2\), which each source provides independently of the other; and synergistic information \(S\), which becomes available only when the sources are used together:
\begin{equation}
I((X_1, X_2); Y) = R + U_1 + U_2 + S
\end{equation}
Thus, when \(X_2\) is added to an existing source \(X_1\), the additional information is:
\begin{equation}
I((X_1, X_2); Y) - I(X_1; Y) = U_2 + S
\end{equation}

Although PID is not the only way to characterize the value of combining information sources, it serves here as an intuitive illustration. It shows that the value of information depends not only on what it provides by itself, but also on which other sources it is combined with. For example, even if two sources individually have almost no predictive power for a target, their joint use may provide crucial information. 

We refer to the connection and recombination of existing information to generate new scientific hypotheses as information collision. By bringing evidence from different sources together, this process may reveal previously unrecognized relationships or suggest explanations that individual records alone do not readily support. A core task of ESI is to identify scientifically meaningful combinations and develop them into hypotheses with testable consequences.

Expansion microscopy provides a clear example of such recombination. Optical microscopes cannot distinguish structures that are too close together, while certain polymer networks expand when placed in water. Spatial separation connects these two ideas: expanding a specimen while preserving the arrangement of its labels could make previously unresolved structures visible. Chen et al. implemented this approach by anchoring fluorescent labels to a swellable polymer network within biological specimens, achieving an effective lateral resolution of approximately $70\,\mathrm{nm}$ with a conventional confocal microscope~\citep{chen2015expansion}.

Both the search for potential collisions and the recombination of information should be guided by key patterns, rather than by every detail of complete records. We define information interfaces as explicit or implicit patterns that allow scientific information from different sources to be connected and recombined. Examples include shared variables, similar mechanisms, and structural analogies across domains. 

Explicit information interfaces can be expressed through structured fields and rules. For example, shared objects and variables help systems match records, while applicability conditions and physical constraints help determine whether those records can be combined. For executable methods, specifying what each operation requires and produces helps agents determine whether different steps can be linked.

Implicit interfaces, in contrast, need to be identified through representations learned by models. Even when two scientific concepts come from different domains and appear unrelated, they may share similar relational structures or underlying connections~\citep{gentner1983structure}. Previous research has explored encoding scientific concepts as vectors, searching representation spaces for potential combinations, and learning which scientific concepts to combine~\citep{sternlicht2026chimera}.

LLMs can participate in both interface identification and subsequent information collisions. Human researchers draw connections among information from accumulated knowledge and insight. LLMs could use their broad pretrained knowledge and learned patterns to explore many such connections across fields, making information collision a systematic research process that can operate continuously across many research questions. Given a research question, a model can draw on papers and knowledge graphs to identify potentially related scientific concepts, then attempt to combine them into new explanations or research proposals~\citep{borrego2025research,ghafarollahi2025sciagents}. Such combinations can also cross disciplinary boundaries. For example, researchers in another field may already have solved a structurally similar problem, allowing the model to adapt their methods to address the current question~\citep{shen2026unlocking}.

The significance of information collision lies not only in the hypothesis it produces, but also in what that hypothesis can set in motion. Models can use information interfaces to connect records and develop testable proposals; experiments then turn these proposals into new evidence, including evidence that rejects them. Once preserved in scientific memory, these results may be recombined with still other observations and trigger new inquiries elsewhere. At the scale of a distributed laboratory network, such sequences could branch and propagate as information chain reactions, in which one experiment changes what is worth testing at multiple other nodes. Multi-agent refinement can extend hypothesis generation~\citep{gottweis2026accelerating}, but the direction and continuity of these chains must ultimately be constrained by empirical evidence. The next challenge is therefore not only what hypotheses to generate, but which claims to test and which results to trust.

\section{What to Test, What to Trust}\label{sec4}

As the preceding sections suggest, accessible, well-organized scientific memory and effective information collision can help models generate more valuable hypotheses. However, scientific intelligence cannot advance through hypothesis generation alone. Without adequate experimental verification, greater computational capacity and richer scientific memory may lead to an accumulation of insufficiently tested hypotheses. Progress therefore depends on coupling the use of accumulated evidence with the acquisition of new, independent evidence to test predictions and guide subsequent exploration.

In physical laboratories, hard constraints arise from sample consumption and instrumental measurements, as well as the time required for reactions and cultivation. Hypotheses can be proposed and developed rapidly, but high-fidelity experimental testing remains expensive and limited in throughput. ESI must therefore decide how to allocate this limited capacity, considering both the potential value of the evidence sought and the reliability of the methods used to obtain it: What experimental information is worth acquiring next, and how much is it worth investing to obtain it?

\subsection{The Value of the Next Experiment}

The value of an experiment can be understood in terms of two objectives: improving performance and reducing uncertainty. Bayesian optimization (BO) and active learning (AL) use explicit acquisition functions to define these two forms of value. BO aims to find the optimum of an objective function. A surrogate model predicts the performance of different candidates and the uncertainty associated with those predictions, while an acquisition function determines which candidate to test next. BO has been widely used to optimize specific scientific objectives, such as identifying optimal experimental conditions, including temperature, concentration, and catalysts~\citep{shields2021bayesian}.

However, the next experiment need not maximize performance. Active learning may instead select experiments to maximize information gain and thereby reduce uncertainty, for example by distinguishing between competing hypotheses~\citep{mackay1992information}.

Both BO and AL are mature and widely adopted methods, yet their application to scientific exploration still has limitations. First, these methods are typically applied to predefined research tasks with specified optimization objectives or acquisition criteria. Second, although previous studies have attempted to transfer knowledge from past tasks to new BO problems, their effectiveness still depends on how historical data are represented and how similar the old and new tasks are~\citep{desimpel2026bayesian}. How to adaptively switch between acquisition strategies and systematically reuse diverse evidence accumulated across studies in future research remains an open challenge.

Performance optimization and uncertainty reduction can also reinforce each other. For example, Entropy Search prioritizes evaluations that are expected to be most informative about the location of the optimum~\citep{hennig2012entropy}. In scientific applications, CAMEO first uses measurements to construct a phase diagram, then searches for materials with optimal target properties, selecting subsequent measurements based on the results obtained so far~\citep{kusne2020fly}. EVOLVEpro represents protein sequences using a protein language model (PLM) and trains a regression model on these representations using a small amount of experimental data. It then iteratively selects mutants with high predicted activity for experimental testing and updates the model with the results, enabling rapid discovery of high-performing proteins through low-N active learning~\citep{jiang2024rapid}. These studies expand the available strategies for exploration, but the challenge of continually applying information acquired from different tasks to new research remains unresolved.

\subsection{The Hidden Cost of Cheap Feedback}

This raises a broader question: Can candidate generation and selection move beyond predefined models and acquisition strategies? Large language models (LLMs) offer another possibility. By drawing on pretrained knowledge and continuously updated experimental records, LLMs can make preliminary assessments of candidates and propose subsequent exploration strategies~\citep{liu2024large}. Functionally, they can participate in tasks traditionally handled by surrogate models and acquisition functions, while offering greater flexibility across tasks and domains. In AlphaEvolve, for example, automated evaluators assess code variants generated by an LLM from existing programs. Promising variants then enter the population for further evolution~\citep{novikov2025alphaevolve,romera2024mathematical}.

However, when this evolutionary search approach is applied to physical experiments, testing a new proposal usually takes much longer than generating it, and scoring criteria as clearly defined as those used to evaluate programs are often unavailable. These difficulties limit the speed and continuity of iterative search. We believe that, as LLMs become increasingly capable of generating and selecting candidates, the bottleneck of scientific discovery shifts toward evaluation and feedback. The critical questions are how to verify each candidate and what feedback should guide the next iteration.

Computational simulations and models can provide preliminary and relatively fast assessments, often helping to reduce costs, but they can also produce false positives and false negatives. False positives consume experimental budgets, while false negatives may prevent genuinely valuable candidates from receiving further testing. Particularly when computational methods cannot reliably predict the target properties, adding another round of inexpensive screening may not save resources. Research on multi-fidelity BO has already shown that unreliable low-fidelity information can actually increase the overall cost of optimization~\citep{mikkola2023multi}.

Moreover, these computational assessments may exert a cumulative influence on subsequent exploration. If a low simulation score is recorded as evidence that a candidate has failed, LLMs may repeatedly encounter this judgment in future iterations. Repeated exposure can reinforce this untested judgment and bias subsequent decisions against the candidate, even though it has never been tested experimentally. In this way, the biases of earlier evaluators may be amplified over time, influencing which candidates are selected for further exploration.

As models and computational capabilities improve, regenerating these intermediate results may become cheaper, and new predictions may be more reliable than older ones. Taken together, these considerations lead us to argue that, in domains where computational methods are not yet sufficiently mature, such readily reproducible intermediate results should not enter long-term scientific memory by default. However, the outputs that informed decisions, together with evaluator versions and provenance, should be retained.

Therefore, computational screening should neither be blindly trusted nor treated as a mandatory first step. When foundation models make hypothesis generation cheap, reliable physical evaluation becomes the true scarcity of discovery—and ESI must integrate candidate evolution with the strategic allocation of that scarce capacity.

\subsection{Evaluation as a Decision}

Rather than require every candidate to follow a fixed screening sequence, ESI should choose the evaluator together with the candidate. Here, an evaluator is defined as a method used to assess a candidate or hypothesis. 

The evaluation action pool covers fact and constraint checking, computational prediction and simulation, and physical experiments. The choice of evaluator should reflect both the question to be resolved and the candidate's potential scientific value. We envisage LLMs taking over the role of explicitly defined acquisition functions in BO and AL by learning which candidates warrant further evaluation. Their assessments of candidate potential would draw on prior scientific knowledge and be continually refined through training on reliable experimental feedback. Candidates judged to offer little scientific value may be deferred or receive only a low-cost assessment, whereas those with substantial potential for performance improvement or uncertainty reduction may justify more expensive testing. 

The selected method should also address what remains uncertain about the candidate. For example, a candidate with promising predicted performance may still require an experimental test of whether it can be prepared. Multi-fidelity optimization provides a concrete basis for joint selection: McDonald et al. selected both molecules and the methods used to evaluate them, considering cost and data quality~\citep{mcdonald2025bayesian}.

When reliable simulation is unavailable, candidates can proceed directly to physical testing. In chemical reaction optimization, Shields et al. updated a BO model using experimentally measured yields without requiring an accurate simulation of the reaction process~\citep{shields2021bayesian}. When suitable computational models are available, extensive optimization can instead precede physical testing. Piggott et al., for example, optimized integrated photonic devices through electromagnetic simulations before fabrication and measurement~\citep{piggott2017fabrication}.

Experimental results can also improve the evaluator itself. If a predictive evaluator repeatedly overestimates the performance of a particular class of candidates, measurements from that class can be used to recalibrate its predictions. The revised evaluator should then be tested against experimental results not used for its update. Once adopted, it can change how subsequent candidates are assessed and selected. Janus provides an example of evaluator evolution in program search, where candidate programs and surrogate evaluators co-evolve, with the latter calibrated through costly computational evaluations~\citep{liu2026janus}.

Both evaluator re-selection and evolution should be grounded in independent experimental evidence rather than agreement within the same generation-and-evaluation loop~\citep{huang2024large,panickssery2024llm}. However, experimental measurements themselves differ in reliability and relevance to the hypothesis being tested~\citep{bertil2014fitness}. These differences should influence how strongly experimental results inform updates to models and evaluators.

\section{The Evolving Scientific Intelligence}\label{sec5}

ESI seeks to connect isolated experimental loops into a network in which discoveries can propagate. Evidence produced at one node may lead another to test a new hypothesis, with the result triggering further investigations elsewhere. Scientific memory and information collision create these opportunities; distributed experimentation determines which connections survive empirical scrutiny. A system capable of sustaining such chain reactions must also revise its evaluators and research strategies when feedback exposes their limitations. These requirements motivate the architecture developed below.

Several key capabilities of scientific intelligence have already emerged. Paper2Agent transforms methods described in research papers into callable tools for agents~\citep{miao2026reimagining}. SciAgents and Co-Scientist demonstrate knowledge-graph-supported hypothesis generation and the iterative refinement of research hypotheses by multiple agents, respectively~\citep{ghafarollahi2025sciagents,gottweis2026accelerating}. Robin connects hypothesis generation with the analysis of real experimental data: researchers perform the experiments, and the system uses the returned results to adjust subsequent research directions~\citep{ghareeb2026multi}. 

\begin{figure}[htbp]
    \centering
    \includegraphics[width=\textwidth]{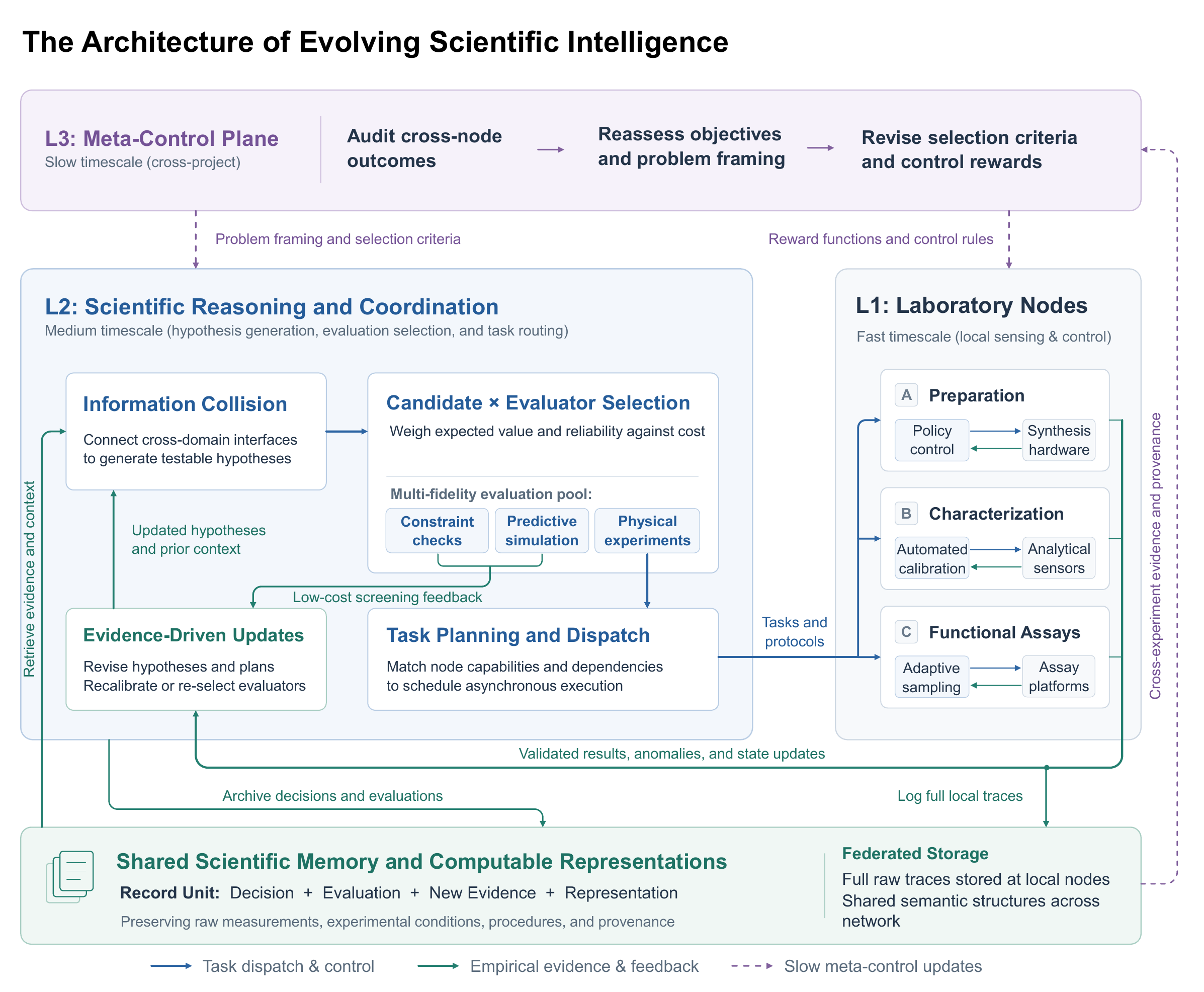}
    \caption
    {
    \textbf{The Architecture of Evolving Scientific Intelligence.}
     ESI enables accumulated evidence to reshape discovery through shared scientific memory and distributed experimentation. L2 serves as the computational hub, generating hypotheses through cross-domain information collision and jointly selecting candidates and evaluators. It coordinates experiments across heterogeneous L1 nodes, where local sensing and control support physical execution. Experimental feedback returns to L2 to revise hypotheses and evaluation methods, while cross-node comparisons help expose setup-dependent biases. Shared memory preserves decisions and evidence with their context and provenance, enabling reuse while retaining full raw traces locally. L3 spans both levels as a slower meta-control plane, reassessing research objectives and revising decision criteria and control rewards to keep optimization aligned with scientific progress. Blue arrows indicate task dispatch and control, while teal arrows indicate evidence and feedback. Dashed purple arrows denote slow meta-control.
    } 
    \label{fig: main}
\end{figure}

These systems offer insights into the design of a self-evolving discovery architecture. Drawing on these insights, ESI organizes scientific capabilities around shared scientific memory, allowing experimental experience to guide revisions to the methods governing subsequent discovery. To coordinate activities that differ in function and timescale, we organize ESI into three levels, as shown in Fig.~\ref{fig: main}. L1 handles local sensing and execution. L2 serves as the main computational hub, generating hypotheses and jointly selecting candidates and evaluators. It also assigns experimental tasks to L1 and coordinates their execution, while revising its methods as evidence accumulates. L3 spans both levels as a slower meta-control plane that reassesses research objectives and decision criteria.

\subsection{Closing the Loop with the Physical World}

An experimental unit can be a scientist or an experimental robot. In the following discussion, we assume that the experimental units are robots. Experimental units perform sensing and operations locally, freeing higher-level research planning from step-by-step equipment control. Once an experimental objective has been set, local controllers or pretrained policy networks execute the task based on on-site observations and promptly correct deviations. The same experimental unit can continuously produce data using established procedures or undertake tasks to test specific hypotheses.

Capability interfaces enable this division of responsibilities. Each node provides machine-readable descriptions of its equipment operations, including the inputs and conditions required for execution. It also tracks equipment and sample states while keeping information on consumable availability and environmental conditions up to date. Capability descriptions tell the coordinator “what this node can do,” while its current state indicates “whether it can do so now.” Existing research provides foundations for these capabilities. SiLA 2 provides standardized communication and functional descriptions for laboratory equipment. SayCan demonstrates how a language model’s assessment of a task can be combined with estimates of robot skill feasibility to select actions that can be performed in the current state~\citep{juchli2022sila,brohan2023can}. After L2 coordinates tasks using this information, the receiving nodes at L1 check the execution conditions and report any missing preparatory steps or resource requirements to the coordinator.

Local control loops can operate on much shorter timescales than scientific planning. We therefore envisage using reinforcement learning policies and other control strategies for real-time regulation within experimental units. For example, a tokamak study deployed a reinforcement learning policy trained in simulation on a physical device, using sensor feedback to control the plasma at 10 kHz~\citep{degrave2022magnetic}. The higher level specifies the target shape and operating requirements, while the local policy continuously assesses the state and controls the system.

In practice, a controller’s ability to detect deviations and respond promptly depends on how sensor measurements are acquired and processed into usable state estimates. Experimental units therefore require both operational and sensing capabilities. Sensing involves identifying and representing key physical states as feedback that the controller can use, allowing the consequences of each action to enter the next observation cycle promptly. Raw measurements and execution histories, along with anomaly records, are preserved alongside experimental results, allowing local experience to feed back into scientific memory and support subsequent policy improvement.

\subsection{Discovery Beyond the Single Laboratory}

L2 carries out the main computational work of ESI. Within project objectives and budgets, coordinating agents at L2 draw on scientific memory to formulate or revise research questions and jointly select candidates and evaluation methods. Information collision at L2 can generate new research directions or new explanations. As new results arrive, the coordinator reassesses which explanations warrant further exploration and adjusts experimental plans. When agents propose conflicting approaches, the system retains the evidence supporting each approach and the key points of disagreement, prioritizing experiments that can distinguish between the explanations or allocating the budget to parallel research branches.

ESI allows nodes to retain different equipment and ways of working. For example, two nodes may bring different experimental capabilities to a shared investigation. One node may prepare a material and perform initial characterization, then share the raw measurements with details of its structure and preparation. Another node can then use its specialized capabilities to address the remaining questions through additional measurements or functional validation. Previous work has demonstrated collaborative reaction optimization across geographically distributed laboratories with similar experimental capabilities~\citep{bai2024dynamic}. ESI seeks to extend such collaboration across heterogeneous experimental capabilities and a broader range of research tasks.

Task decomposition and resource scheduling translate research decisions into work at individual nodes. The coordinator assigns subtasks according to node capabilities and current states, then adjusts responsibilities in response to execution feedback. This mechanism has been explored in heterogeneous multi-robot systems~\citep{liu2025coherent}. For experimental workflows, task dependencies and resource availability determine which steps can begin, while workflow management systems allocate equipment and sample resources~\citep{fei2024alabos}. Different branches can therefore advance asynchronously: while one node waits for a culture to grow or a reaction to finish, others continue experiments or analyze available data~\citep{strieth2024delocalized}.

Distributed operation also requires decisions about which information enters the coordination process and when. Code execution with Model Context Protocol (MCP) illustrates how information can be selectively passed to a model~\citep{jones2025code}. ESI can extend it to experimental networks: nodes retain complete records locally and prioritize reporting results, anomalies, and state changes that may alter experimental plans, while coordinators access detailed data as needed.

Reporting times depend on the impact of results on current decisions. Routine progress can be reported in summaries, whereas information that calls for a change in plans should be communicated promptly. Complete records serve long-term scientific memory; selective communication supports efficient coordination of current tasks.

\subsection{Evolving the Rules of Discovery}

A scientific system can become increasingly efficient at pursuing the wrong objectives. ESI therefore couples rapid execution with slower metacognitive reflection. L1 responds to immediate physical feedback, while L2 revises experimental plans and evaluation methods as research progresses. L3 operates on a longer timescale, drawing on evidence accumulated across experiments to reassess the objectives and decision criteria guiding both.

Consider a policy controlling a device at L1. More aggressive actions may earn higher rewards for speed while also increasing sample loss and the need for repeat measurements. After aggregating these outcomes, L3 can revise how the reward function accounts for speed, loss, and repeat measurements, then retrain and compare policies. Eureka demonstrated this mechanism in simulated robotics tasks: an LLM generated reward functions for policy training, then revised them using task-performance feedback~\citep{ma2024eureka}. To assess these updates, existing and new policies can be compared using common criteria for experimental performance. New policies can first be deployed on a limited set of tasks, with their use expanded as actual improvements are demonstrated.

In our framework, L2 can autonomously revise its evaluation rules. Yet accumulated context and memory may reinforce its commitment to a particular objective even as it improves its methods. L3 periodically steps back from this ongoing optimization to examine how the problem is framed and whether it is still worth pursuing. When accumulated evidence challenges the current direction, L3 can prompt L2 to reconsider the objective and explore alternatives.

Rule updates must be traceable to the experimental records on which they rely; when upstream records are corrected, the associated judgments must also be re-examined. For instance, if a reagent batch or analytical step is found to be problematic, the system can identify which judgments rely on the affected records and arrange further testing.

The broader the impact of a rule change, the more extensive its empirical testing should be. Updates affecting multiple nodes can first be cross-checked by nodes using different equipment, reagent sources, or measurement methods, deliberately varying potential sources of shared error. The system can also periodically conduct experimental checks on key assumptions that support many downstream decisions, directing verification resources toward issues most likely to change the overall research direction.

The tension between fast action and slow reflection is central to this architecture: physical control cannot wait for a comprehensive review of research strategy, while strategic revision requires a broader view than the latest result alone. Separating these timescales lets execution and research planning continue while L3 reassesses what the system should optimize and why.

\subsection{From Local Collaboration to a Global Network}

Large-scale distributed collaboration requires credit attribution mechanisms that participants collectively accept. Research papers provide a widely accepted means of recognizing contributions. For our system, attribution mechanisms for distributed collaboration still need to emerge through practice. In the meantime, ESI can operate through voluntary projects, with participants negotiating recognition according to their actual roles and the progress of the research.

Drawing on open drug discovery practices~\citep{williamson2016open}, this GitHub-like approach to research organizes open collaboration around specific projects. Project organizers and those responsible for experimental nodes negotiate recognition based on their actual roles. To inform these discussions, the system documents experimental work, including method changes and negative results. It also tracks verification activities and the subsequent use of research outputs. Costs and deliverables can be agreed upon before a task begins, while the recognition of scientific contributions is determined jointly as the research progresses.

Under normal conditions, projects can pursue their own research and collaborate selectively. When a shared goal emerges, for example in response to a major public health emergency, funders and project organizers can pool funding and make tasks and experimental resources available to expand parallel exploration. COVID Moonshot demonstrated open, cross-institutional collaboration around an urgent scientific objective~\citep{boby2023open}. ESI’s shared interfaces and record-keeping mechanisms can provide a basis for such temporary mobilization.

In the future we envision, collaboration initiated through early funding can be continuously incentivized by endogenous network effects. When participation in the network offers greater net benefits than independent exploration, distributed collaboration gains the momentum to expand on its own. In a network where information collision and autonomous verification reinforce one another, local discoveries can spread rapidly and trigger new experiments around the world. Scientific exploration may thus transcend local boundaries, becoming a process in which distributed intelligence evolves through the collective efforts of humanity as a whole.

\section{Discussion}\label{sec6}
For discussion, we illustrate ESI with a hypothetical example and compare it with existing frameworks. We then examine the benefits of cross-node collaboration and the challenges that remain.

\subsection{A Cross-Domain Discovery Loop}

To illustrate the workflow of ESI, we consider a hypothetical scenario connecting battery electrochemistry with cell-free enzyme-electrode research. Quinones have been used in both aqueous flow batteries and enzyme electrodes. Their reversible redox behavior and ability to exchange electrons with enzymes provide a chemical basis for this cross-domain connection~\citep{huskinson2014metal,gong2022glucose,mtemeri2023model}.

At Node A, an automated electrochemistry laboratory screens materials for aqueous flow batteries. A quinone candidate Q exhibits stable, reversible redox behavior at low concentrations, but its solubility falls short of the project’s energy-storage requirements and it is excluded from further screening. L1 preserves its raw electrochemical curves, preparation methods, and measurement conditions, together with the reason for rejection. Meanwhile, Node B is developing an enzyme electrode based on FAD-dependent glucose dehydrogenase for repeated glucose measurements. A variant E performs well in a solution-based activity assay but produces a weak electrode response. Changing the mediator also changes the ranking of enzyme variants. Records from both nodes enter shared scientific memory in the form of “Decision + Evaluation + New Evidence + Representation (if available)”, allowing models to access the results and the processes that produced them.

L2 first identifies a distinction in Node A’s records: Q was rejected because of the concentration required for energy storage, while its ability to repeatedly accept and release electrons remains potentially useful. It then connects this property to Node B’s experiments. B needs a mediator that cycles electrons between the enzyme and the electrode, with concentration requirements different from those of a battery. The two sets of records are connected through an explicit information interface: both applications rely on reversible electron transfer. Node A uses this process for energy storage. Node B needs it to carry electrons from E to the electrode. Combining this connection with the observation that E’s performance changes with the mediator, the model proposes a new pairing hypothesis: Q may exchange electrons with E more effectively than the current mediator and thereby improve the electrode response. Information collision thus produces a concrete proposal: test the Q–E pairing, rather than continue screening battery molecules and enzyme variants separately.

L2 focuses evaluation on the unresolved questions surrounding this pairing. Node A checks whether Q can reach the concentration required by B. It then remeasures Q's stability and redox behavior in B's buffer. Node B prepares the enzyme variants and control electrodes in parallel. Once the samples arrive, B measures enzyme–mediator electron exchange and electrode response. It uses enzyme-free and substrate-free controls to assess background current, while independent measurements of substrate consumption or product formation help confirm the target reaction. Researchers perform preparation and transfer steps that have not yet been automated.

If the Q–E pairing shows improved electron exchange and reaction performance under the same enzyme loading and measurement conditions, L2 redirects the next round toward joint optimization of the enzyme and mediator. Meanwhile, L3 may identify a pattern across successive rounds: initial currents keep increasing, but glucose estimates become less accurate during repeated measurements of known standards. In other words, the project has begun to optimize signal magnitude at the expense of reliable glucose measurement. L3 therefore shifts the objective from maximizing initial current to maintaining measurement accuracy over repeated use. In response, L2 repeatedly tests the electrodes with samples of known glucose concentration and selects enzyme–mediator combinations that meet this objective. 

Crucially, the process does not end with a single bilateral transfer; it scales into a multi-node chain reaction. During repeated glucose cycling at Node B, spectroelectrochemical traces reveal that a specific substituent on quinone Q prevents radical dimerization in neutral aqueous buffer—the exact degradation pathway that had secretly limited the lifetime of a related class of organic flow-battery electrolytes. Recorded as a new four-part unit in scientific memory, this finding automatically triggers a reverse collision at Node A to redesign long-lived battery electrolytes, while simultaneously matching an information interface at a third laboratory (Node C) seeking biocompatible redox buffers for cell-free biosynthesis. Moreover, because the provenance graph records that Node A's initially rejected record was essential to Node B's validated sensor and Node C's assay, downstream adoption routes credit and compensation back to Node A, incentivizing further sharing of negative trials. Scaling science in ESI thus means that a single experiment no longer serves a single predefined question: each validated result—positive or negative—multiplies across the network, generating new questions, calibrating shared evaluators, and compounding the return on every unit of experimental effort.

\subsection{Representative Related Works}

Recent advances in AI for Research (AI4R) have begun moving beyond isolated task automation toward integrated discovery frameworks. To clarify the architectural contribution of ESI, we compare our framework with two representative approaches closely related to cumulative discovery: Bridging Literature, Agents, and Zero-gap Experimentation (BLAZE)~\citep{yao2026towards} and Evolutionary Intelligence (EI) for scientific discovery~\citep{wang23evolutionary}.

BLAZE tracks the progress of a project through a shared Research Object, relying on quality gates and human review to guide revisions. While this record shows why a candidate was chosen, the framework does not specify how discrepancies between computational scores and physical measurements should trigger reassessment and revision of the evaluator itself. ESI addresses this gap at L2 by choosing candidates together with their evaluation methods, and then using physical results to recalibrate the evaluators. At a slower timescale, L3 checks whether higher scores under these evaluators actually move the project closer to its real scientific goal.

EI, meanwhile, focuses on retaining past search experience so that learned models and skills can be reused in new tasks. However, reusing tools from previous tasks does not explain how observations collected for completely different purposes can be combined to raise a new scientific question. ESI makes this step explicit: models use information interfaces to spot shared variables or mechanisms across separate records, and then recombine them through information collision into testable hypotheses. This allows evidence from unrelated studies to spark research questions that neither study originally set out to answer.

\subsection{The Network Advantage over Standalone Systems}

Connecting automated laboratories into a distributed network offers clear advantages over running single-node loops in isolation. First, cross-node sharing can help models identify important factors they had not previously considered. Local optimization usually varies selected parameters while treating other conditions as fixed background. When protocols or instruments keep a factor fixed, those experiments provide little evidence of how it affects the outcome. At another node or in another field, however, the same factor may be deliberately varied. Comparing results across conditions can reveal relationships that are difficult to identify within a single experimental setting~\citep{peters2016causal}.

Through cross-domain information collision, LLMs can recognize shared patterns despite differences in terminology and use evidence from one field to reinterpret a problem in another. L2 can then incorporate a previously overlooked factor into the model and design experiments to test its role. Shared information can therefore support performance improvement and reduce uncertainty about the underlying mechanisms. It changes which variables are investigated and which explanations are tested, rather than simply accelerating the optimization of existing parameters.

Second, the network can share laboratories' experience with evaluation methods, not just their equipment. A laboratory may conduct many physical experiments before discovering that a predictor works well for one class of candidates but frequently fails for another. If these findings remain within the original project, other laboratories may have to repeat the same work. ESI can link each evaluator version to the experiments used to assess it, together with the corresponding conditions~\citep{huber2020aiida}. Other nodes can then use these records to judge whether the evaluator is suitable for their candidates. When several nodes work on the same prediction task, their measurements can also be combined to retrain or recalibrate the evaluator. 

Experience from other nodes can also help L2 recognize when a different evaluation method is needed. For example, suppose previous experiments show that high computational scores do not reliably predict cellular activity. A new candidate with a high score would therefore still need to be tested in cells. Instead of continuing to optimize that score, L2 can select a node capable of performing the relevant cell-based assay. In this way, L2 chooses the evaluation method according to the evidence needed, rather than being restricted to the equipment available at a single node. Depending on the task, collaboration may involve groups within one laboratory or teams in different countries.

Third, cross-node feedback can help reveal when a metric keeps improving but the research itself makes no real progress. Within a single-node closed loop, apparent gains may simply reflect measurement biases and evaluator errors compounding across successive iterations. Because each round of candidate selection and model updating depends on the previous round's outputs, a systematic bias in one instrument or simulator is repeatedly fed back into the loop, progressively steering exploration toward local artifacts. Internal replication on the same setup may reproduce the same bias without exposing it. Distributing verification across heterogeneous nodes can interrupt this feedback: setup-specific errors become testable through cross-node comparisons under comparable conditions, giving the network an external basis for reassessing and recalibrating the local evaluator before further optimization reinforces the bias.

A deeper problem emerges when multiple nodes repeatedly encounter the same failure. They improve the chosen metric under different conditions but consistently fail to improve the outcome that matters. This suggests a problem with the assumption that the metric represents the research goal. Improving a surrogate measure does not necessarily improve the final outcome~\citep{vanderweele2013surrogate}.

L3 first reviews the cross-node evidence for shared sources of error. If the gap persists after these checks, L3 can pause optimization under the current metric and direct L2 to test whether changes that raise the metric also improve the intended outcome. L2 designs these experiments and coordinates their execution. If the expected benefit repeatedly fails to materialize, L3 can revise the criterion guiding candidate selection, and L2 replans the research accordingly. Cross-node feedback therefore helps the system correct local measurement errors and reconsider whether its apparent improvements represent genuine progress.

\subsection{Limitations and Outlook}

\subsubsection{Incentives, Attribution, and Market-Driven Valuation}

Process records can trace which data and nodes contributed to a line of inquiry, yet traceability alone does not determine how rewards should be distributed or guarantee that participation is worthwhile for every laboratory. In current scientific practice, researchers have little incentive to share negative results because publications and patents almost exclusively reward positive findings. Simply subsidizing negative results, however, introduces a new dilemma: if a platform assigns static rewards whenever failed trials are uploaded, better-resourced laboratories with higher throughput could capture a disproportionate share of returns, while other participants might be tempted to flood the system with low-effort, random failures.

Overcoming this dilemma requires separating the epistemic utility of a record inside the discovery loop from its external economic valuation. The value of experimental data should be determined not by the act of production itself, but by the downstream scientific demand it helps satisfy. Rather than acting as both platform and central appraiser, ESI can leave pricing to downstream demand while using the model to resolve Arrow's information paradox: a user cannot judge what an unexamined record is worth without seeing it, yet seeing the raw data reveals the information itself. Within ESI, access-controlled records—including negative trials—can participate in L2 information collisions without exposing their raw contents to human users. When a collision produces a promising hypothesis and evaluation plan for a downstream project, the provenance graph records which upstream records were essential to that inference. 
Such provenance and access transactions could be recorded on a blockchain to provide a tamper-evident audit trail across participating laboratories. Smart contracts could then automate royalty distribution along the provenance chain according to agreed attribution and compensation rules. Downstream users who wish to unlock and build upon the resulting protocol then compensate the upstream contributors along this provenance chain, for instance through subscription pools or milestone-based royalties.

Under this demand-driven mechanism, uninformative negative trials rarely enter productive collisions and naturally receive no return, whereas high-value negative records that define critical mechanistic boundaries are rewarded whenever they redirect downstream search. To prevent high-throughput laboratories from monopolizing returns by exhaustively mapping negative spaces, cumulative rewards for a given cluster of redundant negative evidence can be capped at an explicit upper bound—such as a multiple of standardized experimental cost—after which the records transition into the public scientific memory. Designing counterfactual attribution methods that reliably isolate which records were truly decisive in a collision, while keeping transaction friction low for early contributors, remains an important direction for future work.

\subsubsection{Toward a Dynamics of Information Collision}

This paper uses Partial Information Decomposition (PID) to illustrate the potential benefits of combining information, but the collision process and its returns still require formal mathematical treatment. Such a theory must explain not only how two records are combined, but also how models identify which records are worth combining within a massive repository. Information interfaces first help models search for shared scientific patterns, which then guide recombination and reasoning.

Shannon information theory offers a starting point for modeling information gains and losses across this process: for a specified hypothesis space and probability model, it can characterize how new experimental evidence reduces posterior uncertainty. Meanwhile, usable information ($\mathcal{V}$-information) can quantify how much actionable signal a bounded model actually extracts when separate records are recombined, even before new physical measurements are taken. Mechanistic interpretability can further help verify whether models have learned genuine scientific regularities across modalities and disciplines rather than superficial correlations.

Beyond evaluating individual collisions, a deeper challenge is to characterize their network dynamics. As heterogeneous records accumulate, unconstrained recombination risks combinatorial explosion and logical interference—generating more noise than insight. Under what conditions does a single experimental finding trigger a productive chain of new investigations across nodes, how should physical constraints and negative evidence prune unproductive branches, and when do such chains terminate? Answering these questions requires a mathematical framework that governs not only the recombination of scientific information, but also its filtering, propagation, and convergence through experimental evaluation.

\subsubsection{Cross-Domain Representations}

A common data format addresses how information enters the system, whereas cross-domain representation must address how models recognize connections across disparate records. Section 3 proposed sharing semantic structures while retaining encodings suited to specific data types. Yet even after molecules, proteins, and experimental images are encoded, models still need to determine which features can be compared and which relationships can be composed.

We argue that representations for information collision must capture two distinct layers of content. The first is transferable scientific logic, such as ``A regulates C by altering B.'' Different fields may study entirely different objects and use different terminology while sharing the same underlying relational structure. Extracting a finite, reusable set of such relational motifs from scientific descriptions allows models to search for implicit interfaces beyond surface keywords, while still tracking the objects involved and the conditions on which each mechanism depends.

The second layer comprises the specific candidates and their experimental conditions. Relational motifs help models spot candidate connections, whereas detailed multimodal representations help determine whether those connections hold for the physical objects under study. Molecular fingerprints and protein representations from Evolutionary Scale Modeling (ESM) models can be fed into a model together, but the model must learn which features govern their actual physical interaction. As more modalities and domains are introduced, simply concatenating or multiplying vectors does not automatically establish the scientific meaning of the composite representation. Unified scientific tokens face the same challenge: how to make relationships composable across modalities without losing essential domain-specific details.

A complete information unit in ESI therefore links ``Decision + Evaluation + New Evidence + Representation'': why an experiment was selected, how it was evaluated, what outcome was observed, and how that record is encoded for model use. Keeping representations traceable to their underlying records ensures that connections inferred by models can be audited and re-analyzed as representations evolve.

\subsubsection{Reliability and Verification Costs}

Searching for analogous patterns across fields and merging numerical measurements from different laboratories into a single predictive model require different levels of standardization. Cross-domain analogy can generate testable hypotheses by comparing relational mechanisms without requiring identical instruments or a unified measurement scale. Merging numerical data to train a shared predictor, by contrast, requires verifying measurement conditions, aligning physical calibrations, and standardizing units; otherwise, models may mistake instrumental drift or environmental differences for properties of the system under study.

A further challenge is that complete provenance does not guarantee experimental reproducibility. If a receiving laboratory had to re-verify every shared record before using it, verification costs would quickly offset the benefits of data sharing. Within ESI, however, independent replication need not be applied uniformly to all records. Instead, verification can be triggered selectively by downstream demand: when an unverified or anomalous record becomes a critical dependency in a high-value collision, L2 can schedule targeted cross-node replication before committing to costly downstream experiments. How to share the cost of such targeted verification across projects, and how to update node reliability weights based on replication outcomes, remain practical challenges shared across automated laboratory networks.

\subsubsection{Outlook: Scaling Science Through the Information Manhattan Project}

Overcoming the scientific memory wall ultimately requires a paradigm shift in AI for Science: moving from passively mining questions out of existing datasets to actively generating experimental data around unresolved scientific questions. None of the challenges involved in this transition can be solved by a single laboratory or discipline in isolation. We offer ESI not as a finished blueprint, but as a starting point for organizing this collective effort. Realizing it will require domain scientists and AI researchers to work across automated laboratories, building shared scientific memory alongside distributed capacity for experimental evaluation. We call this endeavor the Information Manhattan Project. Its ambition is not merely to scale up isolated models or automate laboratories one by one, but to build an information reactor in which evidence from one investigation can ignite experiments across others. This is how we envision scaling science: through experimentally grounded chain reactions that continually reshape both what we know and how we discover.

\bibliography{sn-bibliography}

\end{document}